\documentclass[journal]{IEEEtran}
\ifCLASSINFOpdf
\else
\fi
\usepackage{multirow}
\usepackage{colortbl}
\usepackage{booktabs}
\newcommand{\etal}{~\textit{et al.}}
\newcommand{\ie}{\textit{i.e.}}

\usepackage{graphicx}
\usepackage{subcaption}
\usepackage{amssymb}
\usepackage{dsfont }
\usepackage{xspace}
\usepackage{amsmath}
\usepackage{cite}
\usepackage[pagebackref=true,breaklinks=true,letterpaper=true,colorlinks,bookmarks=false]{hyperref}

\usepackage[dvipsnames,table,xcdraw]{xcolor}

\begin{document}
%
\title{Long-tailed Recognition by Learning from Latent Categories}

\author{Weide~Liu,
        Zhonghua~Wu,
        Yiming~Wang,
        Henghui~Ding,
        Fayao~Liu,
        Jie~Lin
        and Guosheng~Lin

\thanks{W. Liu is with School of Computer Science and Engineering, Nanyang Technological University (NTU), Singapore 639798 (e-mail: weide001@e.ntu.edu.sg).}

\thanks{Z. Wu is with School of Computer Science and Engineering, Nanyang Technological University (NTU), Singapore 639798 (e-mail: zhonghua001@e.ntu.edu.sg).}

\thanks{Y. Wang is with School of Computer Science and Engineering, Nanyang Technological University (NTU), Singapore 639798 (e-mail:  yiming003@e.ntu.edu.sg).}

\thanks{H. Ding is with School of Electrical and Electronic Engineering, Nanyang Technological University (NTU), Singapore 639798 (e-mail: ding0093@e.ntu.edu.sg).}

\thanks{F. Liu is with the Institute for Infocomm Research (I$^2$R)-Agency for Science, Technology and Research (A*star), Singapore, 138632 (e-mail: Liu$\_$Fayao@i2r.a-star.edu.sg).}

\thanks{J. Lin is with the Institute for Infocomm Research (I$^2$R)-Agency for Science, Technology and Research (A*star), Singapore, 138632 (e-mail: Lin-J@i2r.a-star.edu.sg).}

\thanks{G. Lin is with School of Computer Science and Engineering, Nanyang Technological University (NTU), Singapore 639798 (e-mail: gslin@ntu.edu.sg).}

\thanks{Corresponding author: Guosheng Lin.}
\thanks{W. Liu and Z. Wu make equal contributions.}

}

\markboth{}%
{Shell \MakeLowercase{\textit{et al.}}: Bare Demo of IEEEtran.cls for IEEE Journals}

\maketitle

\begin{abstract}
    In this work, we address the challenging task of long-tailed image recognition. Previous long-tailed recognition methods commonly focus on the data augmentation or re-balancing strategy of the tail classes to give more attention to tail classes during the model training. However, due to the limited training images for tail classes, the diversity of tail class images is still restricted, which results in poor feature representations. In this work, we hypothesize that common latent features among the head and tail classes can be used to give better feature representation. Motivated by this, we introduce a Latent Categories based long-tail Recognition (LCReg) method. Specifically, we propose to learn a set of class-agnostic latent features shared among the head and tail classes. Then, we implicitly enrich the training sample diversity via applying semantic data augmentation to the latent features. Extensive experiments on five long-tailed image recognition datasets demonstrate that our proposed LCReg is able to significantly outperform previous methods and achieve state-of-the-art results.
\end{abstract}


\IEEEpeerreviewmaketitle

\section{Introduction}
\IEEEPARstart{W}{ith} the successful development of Convolution Neural Networks (CNNs), image recognition has achieved great success on the ideally collected balanced datasets (e.g., ImageNet~\cite{imagenet} and Oxford Flowers-102~\cite{flower}). However, in most real-world applications, the natural image data often follows a long-tail distribution, where a few classes have abundant labeled images while most classes have only a few instances or a few annotations. The classification performance of the tail classes on such unbalanced datasets drops quickly with the conventional fully supervised training strategy. 

Long-tailed image recognition has been proposed to address the imbalanced training data problem. The main challenges are the difficulties of handling the small-data learning problems and the extreme imbalanced classification over all the classes. Most of the long-tailed recognition methods~\cite{DBLP:conf/nips/WangRH17,DBLP:journals/pami/HuangLLT20,mikolov2013distributed,article,DBLP:journals/corr/abs-1710-05381,DBLP:conf/eccv/ShenLH16,DBLP:conf/eccv/SarafianosXK18} focus on generating more data samples of tail classes via data augmentation or using the re-balancing strategy to provide higher importance weights for the tail classes. For example, widely used data augmentation techniques like cropping, flipping, and mirroring are used to generate more data samples of the tail classes during the model training. However, we argue that the diversity of the training samples for the tail classes is still inherently limited due to the limited number of training images, which leads to subtle performance improvement for the long-tailed recognition task by using these conventional data augmentation methods.

Different from the conventional data augmentation methods, semantic data augmentation~\cite{isda} tries to augment the image features by adding class-aware perturbations. The perturbations are sampled from the multivariate normal distribution, where the class-wise covariance matrices are calculated from all the training samples.
However, directly applying semantic data augmentation to the long-tailed recognition task may not be optimal. This is because the calculated covariance matrix of the tail classes may not constitute satisfactory meaningful semantic directions for semantic augmentation due to the limited training samples.
MetaSAug~\cite{metasaug} tries to solve the imbalanced statistics problem by updating the class-wise covariance matrix by minimizing the LDAM loss on the validation sets.
However, the performance is still constrained due to the limited diversity and training samples of the tail classes.

\begin{figure*}[t]
\centering
    \includegraphics[width=1\linewidth]{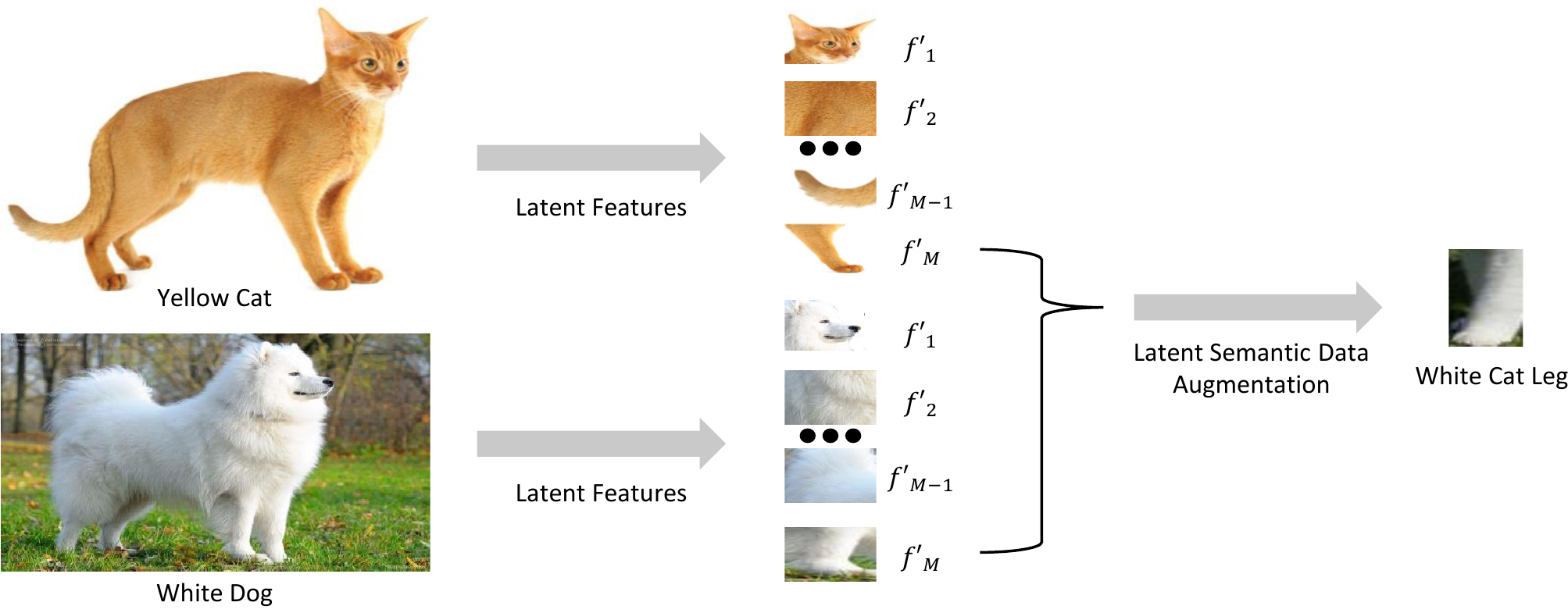}
    \caption{Our LCReg first projects the image features into the latent category features which share the commonality, such as the legs of cats and dogs. By performing the class semantic transformations along with the latent category, we aim to enrich the cat's feature by leveraging the common features, e.g., change the yellow cat leg by leveraging the dog's leg features. 
}
\vspace{1cm}
    \label{Fig:first}
\end{figure*}

To overcome the limitations mentioned above, we propose to mine out and augment the common features among the head and tail classes to increase the diversity of the training samples. The commonality is obtained with the assumption that objects from the same domain might share some commonalities. For instance, cats and dogs share a commonality of legs with similar shapes and appearances. Motivated by this, we argue that it is feasible to re-represent the object features with the common features belonging to the `sub-categories' i.e., each category contains parts of the target objects. For example, as shown in Figure~\ref{Fig:first}, we can re-represent the dog and cat with a series of shared `sub-categories' (e.g., head, leg, body, and tail) with different weights.

\begin{figure*}[t]
\centering
    \includegraphics[width=1\linewidth]{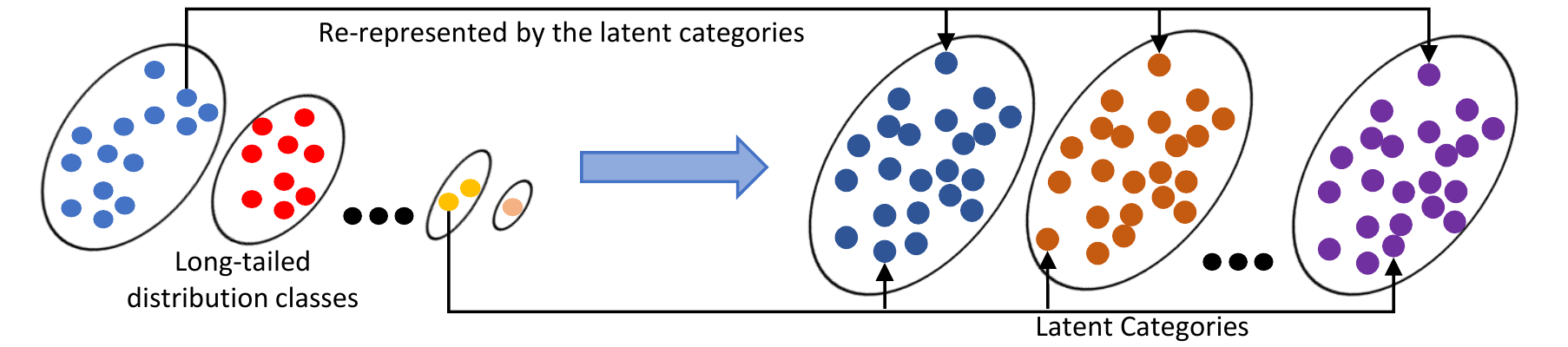}
    \caption{Our LCReg re-represent each object from the original long-tailed distribution dataset by the similarity-weighted sum of latent categories. The latent categories are shareable among the head and tailed classes and form a new balanced distributed dataset. 
    }
    \label{Fig:motivation}
\end{figure*}

Specifically, we introduce a latent feature pool to store the common features, which can be learned through the back-propagation during the model training. As shown in Figure~\ref{Fig:motivation}, the latent features from the pool are class-agnostic and shareable among all the classes. To ensure the latent features are meaningful and sufficient to represent object features, we apply a reconstruction loss to reconstruct the original object features with latent features. Each latent feature contributes to reconstructing the object with a similarity weight. Moreover, to further enrich the diversity of the training data, we implicitly apply a semantic data augmentation method to the latent feature pool. Our method has several advantages with the shareable latent features: 1) We transfer all the object features to the shareable latent categories, making the latent features class-agnostic, which allows our approach to no longer to be constrained to the imbalance distribution. This leads to 2) the tail class objects can benefit from the thriving diversity of the head with the shareable latent features. 3) The tail classes can benefit from the data augmentation technique with the increased diversity, which allows us to develop a latent semantic data augmentation in the latent space. 

The main contributions of this work are concluded from three aspects:
\begin{itemize}
\item We design a Latent categories-based long-tail Recognition (LCReg) method to address the training data imbalance problem. The proposed LCReg explicitly learns the commonalities shared among the head and tail classes for better feature representations.
\vspace{0.35cm}
\item We adopt a semantic data augmentation method on our proposed latent category features to implicitly enrich the diversity of the training samples.
\vspace{0.35cm}
\item We conduct extensive experiments on multiple long-tailed recognition benchmark datasets~(i.e., CIFAR-10-LT, CIFAR-100-LT, ImageNet-LT, iNaturalist 2018, and Places-LT) to validate the effectiveness of our LCReg and achieve state-of-the-art performance.
\end{itemize}

\section{Related Work}
\subsection{Long-Tailed Recognition.}
Imbalanced classification is an extensively studied area ~\cite{salakhutdinov2011learning, zhu2014capturing, bengio2015the, liu2015deep, zhu2016we, ouyang2016factors, liu2016deepfashion, van2017devil, cui2018large,liu2019large, cui2019class, cao2019learning, kang2019decoupling, ye2020identifying, menon2020logit, wu2020distribution, wu2021adversarial, ren2022balanced,tnnls1,tnnls2,liu2020crnet,liu2020guided,liu2020weakly,liu2021cross,liu2021few,liu2021few2,liu2022crcnet,hou2022distilling,hou2022interaction,zhang2020splitting}. Most of the current works can be divided into the re-sampling category: such as the methods of under-sampling head classes, over-sampling tailed classes and data instance re-weighting.

\textbf{Re-sampling and Re-weighting}
Data re-sampling and loss re-weighting are common approaches for long-tailed recognition tasks. The core idea of data re-sampling is to forcibly re-balance the datasets by either under-sampling head classes \cite{DBLP:journals/corr/abs-1710-05381,DBLP:journals/corr/More16a,article} or over-sampling tail classes \cite{DBLP:journals/corr/abs-1710-05381,DBLP:conf/eccv/ShenLH16,DBLP:conf/eccv/SarafianosXK18}. Likewise, loss re-weighting \cite{DBLP:conf/nips/WangRH17,huang2016learning,mikolov2013distributed,DBLP:journals/ida/JapkowiczS02,DBLP:conf/cvpr/TanWLLOYY20} approaches try to balance the loss of semantic classes according to their respective number of samples. However, these re-balancing approaches need careful calibration of weighting to prevent the training from overfitting to tail classes or underfitting to head classes. In particular, the data re-sampling approaches often result in insufficient training of head classes or overfitting to the tail classes; the loss re-weighting approaches suffer from unstable optimization during training \cite{zhong2021improving}. In contrast to the re-sampling and re-weighting methods, our proposed LCReg transfers the unbalanced object features to the shareable and balanced latent categories to learn the commonalities shared among the head and tail classes.

\textbf{Decoupled Training}
The decoupled training scheme \cite{decouple} analyzes and finds that training with the entire long-tailed dataset is beneficial to the feature extractor but harmful to the classifier. Therefore, this two-stage approach proposes first to train the feature extractor and the classifier with the whole long-tailed datasets and then to finetune the classifier with the data re-sampling to balance the weight norm of each semantic class in the classifier. 
The bilateral-branch network equivalently proposes the decoupled training scheme in the same period as \cite{decouple} by adding an extra classifier for the finetuning such that the two-stage training becomes one.
Besides the two-stage training scheme, a causal approach \cite{tang2020long} proposes to learn the long-tail datasets in an end-to-end manner by removing the lousy momentum effect from the causal graph. 
As shown later, our proposed approach is also complementary to the decoupled training.

\subsection{Data Augmentation}
Data augmentation is another line of approaches to facilitate long-tailed recognition, as more augmented samples can alleviate the severely imbalanced distribution of datasets. Recent studies \cite{bbn,zhong2021improving,zhang2020tricks} demonstrate that the mixup helps the tail classes with enriched information from the head classes.
Specifically, \cite{zhong2021improving} additionally proposes label-aware smoothing for finetuning to boost the classification ability. 
Here, we take a step further to explore how the augmentation in the latent category space benefits the long-tailed classification. We follow another type of augmentation called semantic data augmentation \cite{isda} which has been explored recently in domain adaptation \cite{Li2021TSA}. In long-tailed visual recognition, \cite{metasaug} proposes meta-learning to capture category-wise covariance 
for better augmentation. 
Unlike the existing augmentation approaches, we augment the latent category features through a latent semantic augmentation loss to diversify the training samples. We build our proposed method upon \cite{zhong2021improving} to show that our method is also complementary to the data augmentation approaches.
\section{Method}
\begin{figure*}[t]
\centering
    \includegraphics[width=1\linewidth]{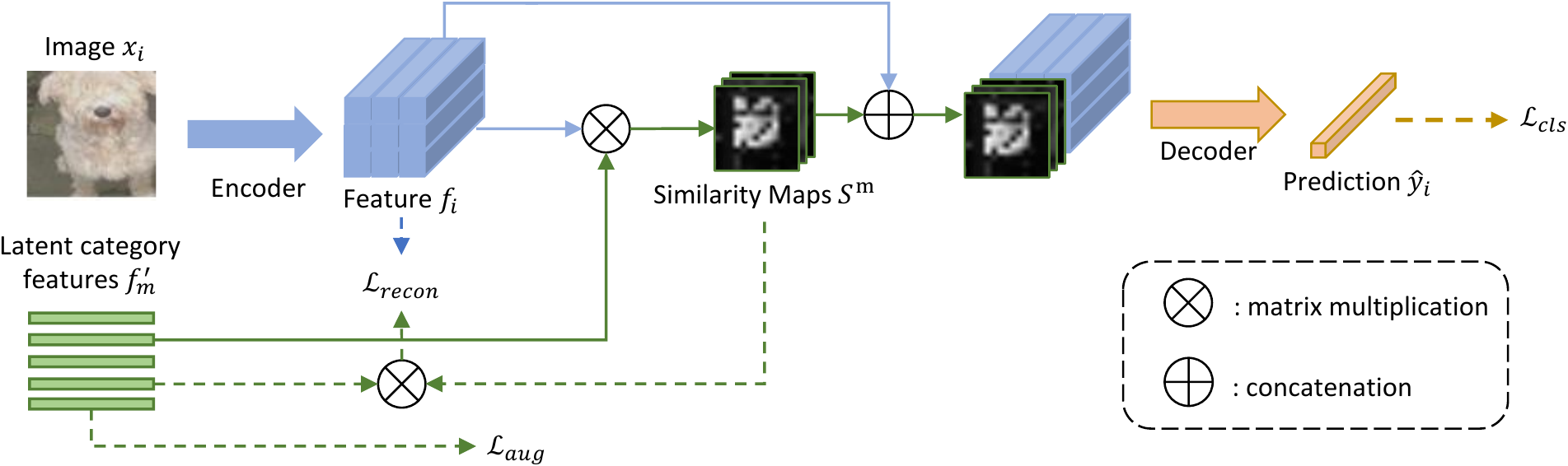}
    \vspace{0.2cm}
    \caption{The pipeline of our proposed LCReg. 
    Given one image, we encode the image features with a set of CNN blocks (such as Resnet 18). By comparing a set of shareable latent category features (each latent category represents part of the object features, initialized by a random learnable embedding, and can be trained through back-propagation.) to the image features, we are able to obtain a set of similarity maps. By applying a latent implicit augmentation loss to the shareable latent category features, we increase the diversity of the latent categories. Combining the similarity maps to the original image features, we predict a final prediction with a decoder. 
    }
    \label{fig:overall}
\end{figure*}

Given that a long-tail distributed dataset contains $N$ training samples with $C$ classes, we sample the $i^{th}$ training sample $x_i$ and its corresponding label $y_i$ from the dataset. The final prediction for $i^{th}$ sample $\hat{y_i}$ is obtained from a classifier using the object feature $f_i \in \mathbb{R}^{D \times H \times W}$, which is generated by the encoder with parameters $\theta$. Our training objective is to optimize the parameters $\theta$ and the classifier to minimize the distance between the prediction $\hat{y_i}$ and the ground truth $y_i$. However, for long-tail distributed datasets, due to the imbalance distribution among each class, most of the features $f_i$ are obtained from the head classes, which makes the classification model biased toward the head classes, resulting in unsatisfactory performance on the tail classes. To alleviate the bias problem, we introduce a set of class-agnostic latent features $f'$, which store the common features among all the classes. In particular, each latent feature contributes to one part of the object features weighted by a similarity score. Moreover, we apply semantic data augmentation to the latent categories to further enrich the diversity of the training samples. The pipeline of our proposed LCReg is shown in Figure~\ref{fig:overall}.

\subsection{Latent category features}
Firstly, we introduce a set of shareable latent features $f'_{0}, f'_{1}, ...f'_{m}, ... f'_{M}$. Each latent feature depicts a latent category representing part of the object features, which is initialized by a random learnable embedding with a dimension of $D$ and can be trained through back-propagation. The shape of each latent feature is $f'_{m} \in \mathbb{R}^{D \times 1}$, so all the latent feature shape is $\mathbb{R}^{D \times M}$.

We further calculate the similarity maps between latent features $f' \in \mathbb{R}^{D \times M}$ and image features $f \in \mathbb{R}^{D \times HW}$ from the image encoder, which benefits the following reconstruction process. 

\begin{equation}
    S^{m} = \sigma( \mathcal{FC}(f'_{m})^T f), 
\end{equation}
where $S^{m} \in \mathbb{R}^{1 \times H \times W}$ indicates the $m_{th}$ similarity map obtained by the $m_{th}$ encoded latent feature $\mathcal{FC}(f'_{m}) \in \mathbb{R}^{D \times 1}$ and the image feature $f$. The $\mathcal{FC}$ is a $1\times1$ convolutional layer to encode the latent features.
We normalize the map with a Sigmoid function $\sigma(\cdot)$ and then reshape the similarity map.

\subsection{Reconstruction Loss}
To encourage the latent features containing more object information, we use the latent features to reconstruct the image features $f$ by employing a reconstruction loss. Specifically, with the similarity maps ${S} \in \mathbb{R}^{M \times H \times W}$ generated by latent features, we apply a Softmax function over all the $M$ similarity maps to identify the most discriminative object parts for each latent category ${S^{m}} \in \mathbb{R}^{1 \times H \times W}$:
\begin{equation}\label{eq:get-smilarity-map}
    \hat{S^{m}} = \frac{\exp(S^{m})}{\sum_{k=1}^{M} \exp(S^{k})}.
\end{equation}

Then we reconstruct image features $f$ by summarizing all the latent categories with the weights from the normalized similarity maps:
\begin{equation} \label{eq:reconstruct feature}
    \hat{f} = \sum_{m=1}^{M} \mathcal{FC}(f'_{m}) \hat{S}^{m} .
\end{equation}

To compare the reconstructed features $\hat{f} \in \mathbb{R}^{D \times HW}$ and the origin features $f \in \mathbb{R}^{D \times HW}$, we calculate the correlation matrix $C_f = \hat{f}^T f$, where $C_f \in \mathbb{R}^{HW \times HW}$ and $H,W$ are the feature size. Finally, we employ a cross-entropy loss to maximize the log-likelihood of the diagonal elements of the correlation matrix $diag(C_f)$ to encourage each latent feature to learn distinct features:

\begin{equation}
    \mathcal{L}_{Recon} = -\sum_{j=1}^{HW}t_j log(\psi(diag(C_f))_j),
\end{equation}
where $j$ is the $j^{th}$ diagonal element of the correlation matrix, and $t_j \in {1,2,...,HW}$ is the pseudo ground truth of the diagonal element, we define the first diagonal element of the correlation matrix $diag(C_f)$ to be the first category, the second one as the second category, and the rest in the same manner. The $\psi(diag(C_f))_j$ denotes the Softmax probability for the $j^{th}$ category.

\begin{table*}[t]
\centering
\resizebox{0.63\linewidth}{!}{
\begin{tabular}{l|l|c|c|c} 
\toprule[1.5pt]
Dataset                           & Methods      & Many   & Medium & Few    \\ \hline

\multirow{2}{*}{CIFAR10-LT IF 100}       & Ours$^*$     & {90.9} & {80.8} & {73.7} \\
                                  & \cellcolor[HTML]{9B9B9B}Ours         & \cellcolor[HTML]{9B9B9B}\textbf{92.6} & \cellcolor[HTML]{9B9B9B}\textbf{81.5} & \cellcolor[HTML]{9B9B9B}\textbf{75.4} \\   \hline

\multirow{6}{*}{CIFAR100-LT IF 100}      & OLTR~\cite{liu2019large}         & 61.8   & 41.4   & 17.6   \\ 
                                  & LDAM + DRW~\cite{cao2019learning}   & 61.5   & 41.7   & 20.2   \\
                                  & ${\tau}$-norm~\cite{decouple}       & 65.7   & 43.6   & 17.3   \\
                                  & cRT~\cite{decouple}          & 64.0   & 44.8   & 18.1   \\
                                  & Ours$^*$     & 63.1       & {48.4}       &  {25.3}      \\
                                  & \cellcolor[HTML]{9B9B9B}Ours         & \cellcolor[HTML]{9B9B9B}\textbf{64.2} & \cellcolor[HTML]{9B9B9B}\textbf{49.2}  & \cellcolor[HTML]{9B9B9B}\textbf{25.3}  \\  \hline
\multirow{6}{*}{ImageNet-LT}      & cRT~\cite{decouple}          & 62.5   & 47.4   & 29.5   \\
                                  & LWS~\cite{decouple}          & 61.8   & 48.6   & 33.5   \\
                                  & Ours$^*$     & 61.7   & 51.3   & 35.8   \\
                                  & \cellcolor[HTML]{9B9B9B}Ours         & \cellcolor[HTML]{9B9B9B}\textbf{66.2}   & \cellcolor[HTML]{9B9B9B}\textbf{52.9}   & \cellcolor[HTML]{9B9B9B}\textbf{35.8}   \\  \hline
\multirow{5}{*}{iNaturalist 2018} & cRT~\cite{decouple}          & 73.2   & 68.8   & 66.1   \\
                                  & ${\tau}$-norm~\cite{decouple} & 71.1   & 68.9   & 69.3   \\
                                  & LWS~\cite{decouple}          & 71.0   & 69.8   & 68.8   \\
                                  & Ours$^*$     & 73.2   & 72.4   & 70.4   \\
                                  & \cellcolor[HTML]{9B9B9B}Ours         & \cellcolor[HTML]{9B9B9B}\textbf{73.8} & \cellcolor[HTML]{9B9B9B}\textbf{73.4} & \cellcolor[HTML]{9B9B9B}\textbf{71.5} \\  \hline
                                  
\end{tabular}
}
\vspace{2mm}
\caption{
We report accuracy on three splits of classes: Many, Medium, and Few. We validate our methods on multiple datasets, including small-scale datasets (CIFAR10-LT, CIFAR100-LT with IF 100) and large-scale datasets (ImageNet-LT, and iNaturalist 2018). 
Ours$^*$ indicates ours baseline (without the latent category features, reconstruction loss $\mathcal{L}_{Recon}$ and latent augmentation loss $\mathcal{L}_{latent\_aug}$).}
\label{many-medium-few}
\end{table*}

\begin{table*}[t]
\centering
\resizebox{0.9\linewidth}{!}{
\begin{tabular}{l|c|c|c|c|c|l|c|c|c|c}
\toprule[1.5pt]
Dataset  & \multicolumn{5}{c}{Number of latent class} &  \multicolumn{1}{|l}{Dataset}         & \multicolumn{4}{|c}{Number of latent class} \\
\hline
         & 20         & 30        & 40        & 50        & 60        &                 & 20            & 60           & 100          & 200          \\  \hline
CIFAR-10-LT  & 81.9       & 82.4      & 83.1      & 82.5      & 79.6      & ImageNet-LT     & 54.5          & 55.0         & 55.3         & 55.2            \\
CIFAR-100-LT & 47.1       & 47.2      & 47.4      & 47.6      & 46.1      & iNaturalist 2018 & -             & 71.6            & 71.6         & 72.6        \\
\bottomrule[1.5pt]
\end{tabular}
}
\vspace{2mm}
\caption{Ablation studies of the effectiveness of the number of latent categories. We conduct the experiments on the small dataset (CIFAR-10-LT and CIFAR-100-LT with IF 100) and large dataset (ImageNet-LT and iNaturalist 2018).
The larger the dataset (more training samples and classes), the more latent categories are needed to represent better performances.
}
\label{Table: abalation-emb}
\end{table*}
\begin{table*}[t]
\centering
\small
\begin{tabular}{c|c|c|c|c|c|c|c|c|c}
\toprule[1.5pt]
\multicolumn{3}{c}{Components}              & \multicolumn{3}{|c}{CIFAR-10-LT}  & \multicolumn{3}{|c}{CIFAR-100-LT} & \multicolumn{1}{|c}{iNaturalist 2018}                \\ \hline
latent category & latent aug & latent recon &100 &50 &10 & 100 &50 &10 & - \\ \hline
                &            &              &   82.1   &85.7 &90.0    &    47.0   &52.3 &63.2             &68.9            \\ 
\checkmark      &            &              &   82.2   &85.8 &90.7    &    47.2   &52.6 &63.9             &69.4            \\ 
\checkmark      & \checkmark &              &   82.5   &86.0 &91.0   &     47.4   &53.0 &64.1            &69.8            \\ 
\checkmark      &            & \checkmark   &   83.0   &86.2 &91.1    &    47.3   &52.5 & 64.0            &70.0            \\ 
\checkmark      &\checkmark  & \checkmark   &   \textbf{83.1}   &\textbf{86.5} &\textbf{91.2}    &    \textbf{47.6}   &\textbf{53.1} & \textbf{64.2}            &\textbf{70.5}              \\
\bottomrule[1.5pt]
\end{tabular} 
\vspace{2mm}
\caption{Ablation studies of each component, including whether utilizing our proposed latent category, latent augmentation loss(latent aug) and latent reconstruction loss (latent recon). We conduct the experiments on the small dataset (CIFAR-10-LT and CIFAR-100-LT with IF 100, 50, and 10) and large dataset (iNaturalist 2018).
The results show that each of our proposed components improves the baseline (without any component).
}
\vspace{+0.2cm}
\label{Table: abalation-losses}
\end{table*}
\begin{table}[t]
\centering
\resizebox{0.9\linewidth}{!}{
\begin{tabular}{l|c|c|c|c|c|c}
\toprule[1.5pt]
  & \multicolumn{3}{c}{CIFAR-10-LT}      & \multicolumn{3}{|c}{CIFAR-100-LT} \\
\hline
      Methods   & 100         & 50        & 10                      & 100         & 50        & 10                  \\  \hline
Baseline & {82.1}   & {85.7} &  {90.0}                  & {47.0}  & {52.3}   &  {63.2}        \\
+ ISDA & {79.8}   & {82.7} &  {87.8}                    & {43.5}  & {47.8}   &  {57.7}         \\
+ $L_{Aug}$    & \textbf{82.5}   & \textbf{86.0}    &\textbf{91.0}   & \textbf{47.4}  & \textbf{53.0}   &  \textbf{64.1}        \\
\bottomrule[1.5pt]
\end{tabular} 
}
\vspace{2mm}
\caption{Ablation studies for the normal feature augmentation with ISDA and the latent feature augmentation. $L_{Aug}$ denotes the method applying the latent augmentation method on the latent category features. We report the top-1 accuracy (\%) on different datasets.
}
\label{Table: abalation-isda}
\end{table}

\subsection{Latent Feature Augmentation}
Data augmentation is a powerful technique that has been widely used in recognition tasks to increase training samples to reduce the over-fitting problem. 
Traditional data augmentation, such as rotation, flipping, and color-changing, are utilized to increase the training samples by changing the image itself.
In contrast to conventional data augmentation techniques, semantic data augmentation augments the semantic features by adding class-wise conditional perturbations~\cite{isda}. 
The performance of such class-conditional semantic augmentation heavily relies on the diversity of the training samples to calculate significant, meaningful co-variance matrices for perturbation sampling. However, in the long-tail recognition task, the diversity of tail classes is low due to the limited training samples. The calculated class-conditional statistics will not include sufficient meaningful semantic direction for feature augmentation, which causes negative effects on long-tailed recognition tasks. The details are shown in Section~\ref{sec. ablation-isda} and Table~\ref{Table: abalation-isda}. 

\textbf{Latent implicit semantic data augmentation.}
In contrast with ISDA~\cite{isda}, we propose to augment the latent categories to implicitly generate more training samples. To implement the semantic augmentation in the latent feature categories directly, we calculate the co-variance matrices ($\boldsymbol{\Sigma}=\{\boldsymbol{\Sigma}_1, \boldsymbol{\Sigma}_2, ...,\boldsymbol{\Sigma}_M\}$) for each latent category by updating the latent features $f'_m$ at each iteration over total $M$ classes. In particular, for the $t^{th}$ training iteration, we have total $n_m^{(t)} = n_m^{(t-1)} + n^{'(t)}_m$ training samples for $m_{th}$ latent category, where the $n^{'(t)}_m$ denotes the number of training samples at the current $t^{th}$ iteration for $m_{th}$ latent category. Then we estimate the average latent feature value $\mu_m^{(t)}$ of $m_{th}$ latent category for total $t$ iteration with:

\begin{equation}
    \label{ave}
    \mu_m^{(t)} = \frac{n_m^{(t-1)}\mu_m^{(t-1)} + n^{'(t)}_m {\mu'}_m^{(t)}}
    {n_m^{t}},
\end{equation}
where the ${\mu'}_m^{(t)} = \frac{1}{n^{'(t)}_m} \sum_{1}^{n^{'(t)}_m}f'_m $ denotes the current average values of the latent $m_{th}$ class features at $t^{th}$ iteration. 
Then we can update the $m_{th}$ latent category covariance matrices for total $t$ training iteration with:
\begin{equation}
    \label{cv}
    \begin{split}
\Sigma_m^{(t)} 
         = \frac{n_m^{(t-1)}\Sigma_m^{(t-1)} + n^{'(t)}_m {\Sigma}_m^{'(t)}}
        {n_m^{(t)}} + \\
        \frac{n_m^{(t-1)} n^{'(t)}_m \Delta(\mu)\Delta(\mu)^T}
        {(n_m^{(t)})^2},
    \end{split}
\end{equation}
where $\Delta(\mu) = (\mu_m^{(t-1)} - {\mu'}_m^{(t)})$, and the ${\Sigma'}_m^{(t)}$ denotes the  $m_{th}$ latent category covariance matrices at current $t^{th}$ iteration.

Then, we augment the features by sampling a semantic transformation perturbation from a Gaussian distribution $\mathcal{N}(0, \lambda\boldsymbol{\Sigma}_{y'_m})$, where $\lambda$ indicates the hyperparameter of the augmentation strength and $y'_m \in {1,...,M}$ indicates the pseudo ground truth of the $M$ latent categories. In particular, we set the first latent category as the first class, the second one as the second class, and the rest in the same manner.
For each augmented latent feature $f^a_m$ we have \begin{equation}
    f^a_m \sim \mathcal{N}(f'_m, \lambda\Sigma_{y'_m}). 
\end{equation}

Furthermore, when we sample infinite times to explore all the possible meaningful perturbations in the $\mathcal{N}(0, \lambda\boldsymbol{\Sigma}_{y'_m})$, there is an upper bound of the cross-entropy loss~\cite{isda} on all the augmented features over $N$ training samples:
\begin{align}
    \label{Eq: loss-embedding}
    \mathcal{L}_{latent\_aug} &= \sum_{i=1}^{N} L_{ \infty}(f(\boldsymbol{x_i};\theta),y'_m; \boldsymbol{\Sigma}) \\ \notag
& = \frac{1}{N} \sum_{i=1}^{N} log(\sum_{j=1}^{M} e^{ z_j
})
\end{align}
\begin{equation}
\begin{split}
z_j = (\boldmath{w}^{T}_{j} - \boldmath{w}^{T}_{y'_{m}})f^a_m + (b_{j} - b_{y'_{m}}) + \\ \frac{\lambda}{2}(\boldmath{w}^{T}_{j} - \boldmath{w}^{T}_{y'_{m}})\Sigma_{y'_m}(\boldmath{w}_{j} - \boldmath{w}_{y'_{m}}),
\end{split}
\end{equation}
where $\theta$ indicates the encoder parameters for the latent category features. 
$\boldmath{w}$ and $\boldmath{b}$ are the weight and biases corresponding to the a $1\times1$ convolution layer $\mathcal{FC}$ motioned above. Following ISDA~\cite{isda}, we let $\lambda  = (t/T)\!\times\!\lambda_0$ to reduce the augmentation impact in the beginning of the training stage, where $T$ indicates the total iteration. 

With the augmented latent category features, we are able to increase the diversity of training samples by reconstructing the augmented latent features back to the image features $f$ with the reconstruction loss $\mathcal{L}_{Recon}$.

\begin{figure*}[t]
\centering
    \includegraphics[width=1\linewidth]{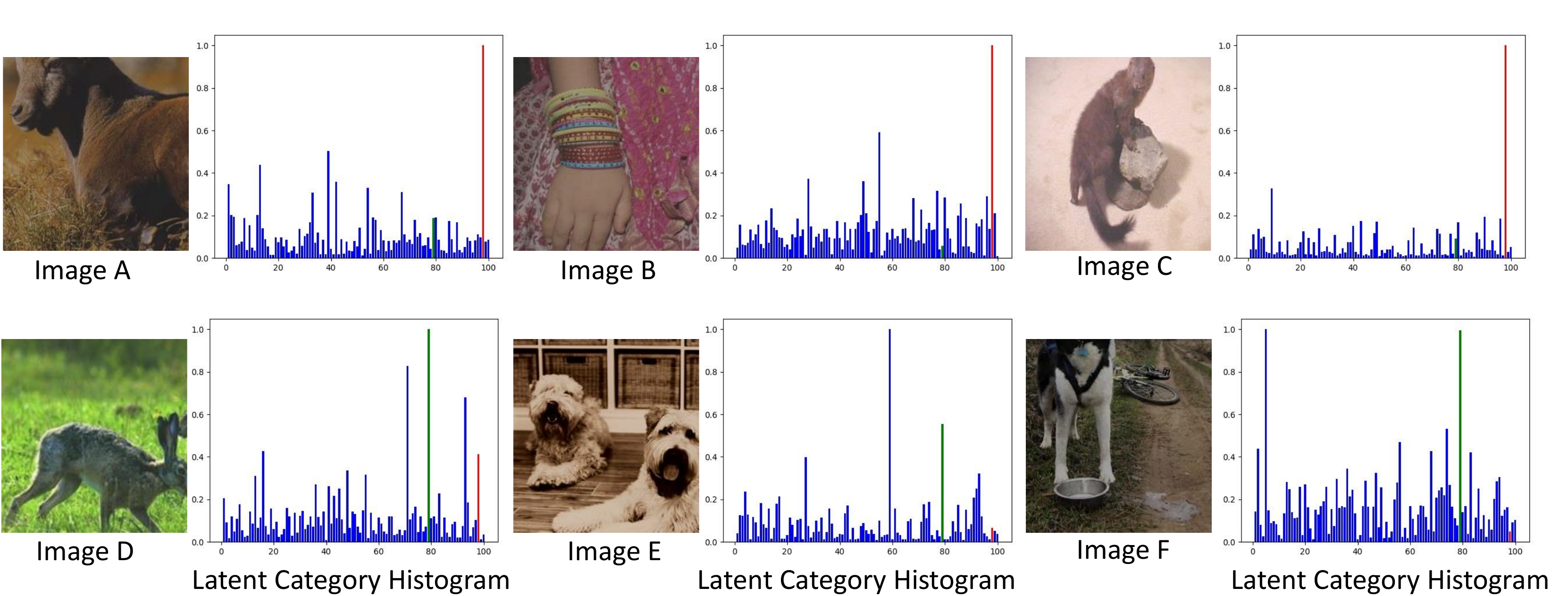}
    \caption{We visualize the weight histogram of latent categories contributing to the reconstruction of the image features. As shown in the figures, the $79^{th}$ latent category (green) is highlighted by the `hare'(Image D), and `dogs'(Image E and F), while all of them contain the similar shape of the limbs. Furthermore, the `cow'(Image A), `human arm'(Image B), and `fisher'(Image C) share some commonalities captured by the $98^{th}$ latent category(red). Note that our proposed method aims to learn some commonalities between images belonging to latent classes. The latent classes are not only denoted for the appearances from the human point of view. A common characteristic can be any characteristic of an object, such as color, structure, or shape.}
    \label{Fig:visualization}
\end{figure*}

\begin{table}[t]
\centering
		\centering
		\resizebox{1\linewidth}{!}{
		\begin{tabular}{l|ccc|ccc}
			\toprule[1.5pt]
			{\multirow{2.6}*{\textbf{Method}}}	 &\multicolumn{3}{c|}{CIFAR-10-LT} & \multicolumn{3}{c}{CIFAR-100-LT} \\ 
			\cmidrule(lr){2-4}\cmidrule(l){5-7}
			& 100 & 50 &  10 & 100 & 50 & 10 \\ 
			\hline
			
			CE (Cross Entropy) & 70.4  & 74.8  & 86.4  & 38.4  & 43.9  &  55.8   \\
			mixup~\cite{mixup} & 73.1  & 77.8 &  87.1  & 39.6  & 45.0  &  58.2   \\
			LDAM+DRW~\cite{ldam} & 77.1  & 81.1  &  88.4  & 42.1 & 46.7  &  58.8    \\

			BBN{\scriptsize{(include mixup)}}~\cite{bbn} & 79.9  & 82.2  &  88.4  & 42.6  & 47.1  &  59.2  \\
			Remix+DRW~\cite{remix} & 79.8  & -  &  89.1  & 46.8  & -  &  61.3  \\

			
			{MiSLAS~\cite{zhong2021improving}}  & {82.1}   & {85.7} &  {90.0}  & {47.0}  & {52.3}   &  {63.2}  \\
			MetaSAug CE\cite{metasaug} & 80.5 & 84.0 & 89.4 & 46.9 & 51.9 & 61.7 \\
			\hline   

			{Ours}  & \textbf{83.1}   & \textbf{86.5} &  \textbf{91.2}  & \textbf{47.6}  & \textbf{53.1}   &  \textbf{64.2} \\ 
			\bottomrule[1.5pt]     
		\end{tabular}
		}
		\vspace{2mm}
		\caption{Top-1 accuracy (\%) for ResNet-32 based models trained on CIFAR-10-LT and CIFAR-100-LT.}
		\label{tab:cifar}
	\end{table}

\begin{table}[t]
\centering
\small
\resizebox{0.8\linewidth}{!}{
\begin{tabular}{p{4.3cm}|p{1.6cm}<{\centering}}
					\toprule[1.5pt]
					\textbf{Method}              & ResNet-50     \\ \hline
					CE & 44.6         \\
					CE+DRW~\cite{ldam} & 48.5         \\
					Focal+DRW~\cite{lin2017focal}  & 47.9 \\
					LDAM+DRW~\cite{ldam}    & 48.8 \\ 
					NCM \cite{decouple}  & 44.3  \\
$\tau$-norm \cite{decouple} & 46.7   \\
cRT \cite{decouple} & 47.3     \\
LWS \cite{decouple} & 47.7     \\
{MiSLAS~\cite{zhong2021improving}}      & {52.7} \\
MetaSAug CE \cite{metasaug} & 47.4\\
						\hline
							Ours  &  \textbf{55.3}      \\
					\bottomrule[1.5pt]
					
\end{tabular}
}
\vspace{2mm}
\caption{Top-1 accuracy (\%) on ImageNet-LT.
}
\label{Table: soa_imagenet}
\end{table}
\begin{table}[t]
\centering
\small
\resizebox{0.8\linewidth}{!}{
\begin{tabular}{p{4.3cm}|p{1.6cm}<{\centering}}
					\toprule[1.5pt]
					\textbf{Method}              & ResNet-50 \\ \hline
					CB-Focal~\cite{effnum}  & 61.1\\
					LDAM+DRW~\cite{ldam}  &68.0 \\ 
					OLTR~\cite{liu2019large}   & 63.9 \\
					cRT~\cite{decouple}  & 65.2 \\
					${\tau}$-norm~\cite{decouple}  & 65.6 \\
					LWS~\cite{decouple} & 65.9 \\
					BBN{\scriptsize(include mixup)}~\cite{bbn} & 69.6 \\
					Remix+DRW~\cite{remix} & 70.5  \\ 
					{MiSLAS~\cite{zhong2021improving}}  & {71.6} \\ 
					MetaSAug CE \cite{metasaug} & 68.8\\
					
					\hline
						Ours  & \textbf{72.6}      \\ 
					\bottomrule[1.5pt]
\end{tabular}
}
\vspace{2mm}
\caption{Top-1 accuracy (\%) on iNaturalist 2018.
}
\label{Table: soa_inat}
\end{table}
\begin{table}[t]
\centering
\small
\resizebox{0.8\linewidth}{!}{
\begin{tabular}{p{4.3cm}|p{1.6cm}<{\centering}}
\toprule[1.5pt]
					\textbf{Method}              & ResNet-152     \\ \hline  
					Range Loss~\cite{zhang2017range}      & 35.1 \\
					FSLwF~\cite{gidaris2018dynamic} & 34.9 \\ 
					OLTR~\cite{liu2019large} & 35.9 \\
					OLTR+LFME~\cite{xiang2020learning} & 36.2 \\
						\hline
					Ours  &    \textbf{40.2}         \\ 
					\bottomrule[1.5pt]
\end{tabular}
}
\vspace{2mm}
\caption{Top-1 accuracy (\%) on Places-LT.
}
\vspace{-0.2cm}
\label{Table: soa_places}
\end{table}

\subsection{Training Process}
We adopt decoupled training for the long-tailed task as in \cite{zhong2021improving}. Specifically, in the first stage of the training process, our training objective includes the reconstruction loss $\mathcal{L}_{Recon}$ which is applied on the latent category features, a latent augmentation loss $\mathcal{L}_{latent\_aug}$ that augments the latent features, and a cross-entropy classification loss which is applied on final prediction $\hat{y_i}$ generated with the decoder.
We optimize the network parameter by combining all the losses:
\begin{equation}
\label{eqn:loss}
\mathcal{L} =\alpha \mathcal{L}_{latent\_aug}  + \beta\mathcal{L}_{Recon} + \gamma\mathcal{L}_{cls}, 
\end{equation}
where ${L}_{cls}$ indicates the final classification loss (CE loss) between the ground truth $y$ and the prediction $\hat{y_i}$.  $\alpha$, $\beta$, and $\gamma$ are the trade-off parameters, which have been set to 0.1, 0.1 and 1, respectively. 
In the second stage of training, following~\cite{zhong2021improving}, we finetune the network. 
\section{Experiments}
\subsection{Implementation Details}
We follow the training pipeline as in ~\cite{zhong2021improving,bbn} to conduct experiments on five datasets, including CIFAR-10-LT, CIFAR-100-LT, ImageNet-LT, iNaturalist 2018, and Places-LT. 
We use the SGD as the optimizer to train the network. We apply data augmentation such as random scale, random crop, and random flip during the training process.
If there is no special declaration, we conduct the experiments with a batch size of 128.

\subsection{Dataset}
\noindent\textbf{CIFAR-10-LT and CIFAR-100-LT.}
Following~\cite{cao2019learning}, we conduct experiments on the long-tail version of CIFAR datasets. CIFAR-10 and CIFAR-100 contain 50000 images and 10000 for training and validation, including 10 and 100 categories, respectively. In particular, we discard the training samples to reorganize a unbalanced dataset with imbalance factor(IF) $\beta = N_{max}/N_{min}$. The $N_{max}$ and $N_{min}$ are the numbers of training samples for the largest and the smallest classes. Following~\cite{cao2019learning,zhong2021improving,bbn}, we conduct the experiments on the CIFAR-LT with imbalance factor(IF) $\beta=10,50$ and $100$.

\noindent\textbf{ImageNet-LT.} Liu~\etal~\cite{liu2019large} propose the ImageNet-LT dataset, which contains 115,846 training images and 50,000 validation images, including 1000 categories, with the imbalance factor(IF) of 1280/5. This dataset is a subset of ImageNet~\cite{imagenet}. They follow the Pareto distribution with power value $\alpha$ = 6 to sample the images and rearrange to a new unbalanced dataset. 

\noindent\textbf{iNaturalist 2018.} iNaturalist 2018~\cite{van2018inaturalist} is a large-scale dataset collected from the real world, whose distribution is extremely unbalanced. It contains 435,713 images for 8142 categories with imbalanced factor(IF) of 1000/2. 

\noindent\textbf{Places-LT.} Places-LT is a long-tailed distribution dataset generated from the large-scale scene classification dataset Places~\cite{places}. It consists of 184.5K images for 365 categories with an imbalanced factor(IF) of 4980/5. 

\subsection{Comparisons with State-of-the-art methods}
\noindent\textbf{Experiments on CIFAR-LT.} Following~\cite{zhong2021improving,tang2020long,ldam,bbn}, we conduct the experiments on CIFAR-10-LT and CIFAR-100-LT with the IF of 10, 50, and 100. The latent categories are set to 40 and 50 for CIFAR-10-LT and CIFAR-100-LT, respectively. As shown in Table~\ref{tab:cifar}, our proposed method outperforms all previous methods.

\noindent\textbf{Experiments on large-scale datasets.} 
We further validate the effectiveness of our method on the large-scale imbalanced datasets, \ie, ImageNet-LT, iNaturalist 2018, and Places-LT. The latent category number is set to 100 for ImageNet-LT and 200 for iNaturalist 2018, and 100 for the Places-LT dataset.
As shown in Table~\ref{Table: soa_imagenet}, Table~\ref{Table: soa_inat} and Table~\ref{Table: soa_places}, our proposed method outperforms all the other methods and achieves the new state-of-the-art performance on all the large-scale datasets.

\subsection{Ablation Studies}
\noindent\textbf{Number of the latent categories.}
We conduct experiments to analyze how the latent categories affect the performance of different datasets. As shown in Table~\ref{Table: abalation-emb}, we experiment on both small and large scale datasets to explore the effectiveness of the number of latent categories. 
For the larger datasets, which contain more training samples and classes, we suggest using more latent categories to represent the original image features to achieve better performances.
However, enlarging the number of latent categories could not continuously increase the performances. For example, 40 categories yield the best performance on the CIFAR-10-LT dataset. Continually increasing the number of categories would drop the performances very quickly. 
We speculate that if there are too many latent categories, each object feature might be split too finely by the latent features, failing to obtain the meaningful parts.

\noindent\textbf{Performance on different splits of classes.}
We further report the classification accuracy for the many (more than 100 images per class), medium (20 to 100 images per class), and the few (less than 20 images per class) classes. In particular, we set the number of latent categories as 40 for CIFAR-10-LT, 50 for CIFAR-100-LT, 100 for ImageNet-LT, and 200 for iNaturalist 2018.
As shown in Table~\ref{many-medium-few}, our method achieves the best performance on the many, medium, and few classes by a large margin for all the datasets. Specifically, on the ImageNet-LT `many' dataset, our LCReg achieves a 4.4\% accuracy gain over the previous SOTA methods while keeping the performances of medium and few classes not dropped. 

\noindent\textbf{Effect of each component.}
We investigate the contribution of each component of our proposed method: the latent categories, the latent augmentation loss, and the latent reconstruction loss. We conduct the ablation experiments on both the small and large scale datasets to validate our method. Specifically, we choose the imbalance factor(IF) to 100 and set the number of latent categories to 40 for CIFAR-10-LT and 50 for CIFAR-100-LT.
For the experiment on the large challenge datasets(iNaturalist 2018), we set the number of latent categories to 100 with a small training batch size(16) due to resource limitations. As shown in Table~\ref{Table: abalation-losses}, only adding our proposed latent categories could have a significant improvement over the baseline method for all the datasets. The performances are further improved by applying the latent augmentation loss and the latent reconstruction loss.

\noindent\textbf{Visualization of the latent categories.}
As shown in Figure~\ref{Fig:visualization}, we visualize the latent category histogram on the ImageNet-LT dataset with 100 latent categories. We reconstruct the image features with the latent categories, and each latent category contributes with a normalized similarity weight generated by equation~\ref{eq:get-smilarity-map}. As shown in the figure, the $79^{th}$ latent category (green) is highlighted by the `hare' and `dogs' (Image E and F), while both of them contain similar limb patterns. Furthermore, the `cow', `human arm', and `fisher' also share some commonalities captured by the $98^{th}$ latent category(red). To be noticed, our proposed method aims to learn some commonalities between the images which belong to the latent classes, not only the appearance from the human perspective. The common features can be any character of the objects, such as the color or the shape.

\noindent\textbf{Latent augmentation vs. ISDA} \label{sec. ablation-isda}
As shown in Table~\ref{Table: abalation-isda}, we directly apply the ISDA~\cite{isda} method to the original class features and conduct the experiments on CIFAR-10-LT and CIFAR-100-LT with different imbalance impacts. 
In contrast to directly using the unbalanced features, our latent augmentation method augments the features among the latent categories. It brings significant improvement to all the long-tail recognition datasets. Specifically, we set the number of latent categories as 40 for CIFAR-10-LT and 50 for CIFAR-100-LT.

\section{Conclusion}
In this paper, we have proposed a latent category-based long-tail recognition(LCReg) method to increase the diversity of the training samples for long-tailed recognition tasks by mining out the common features among the head and tail classes. We adopt a semantic data augmentation method on our proposed latent category features to implicitly enrich the diversity of the training samples. Experiments on several long-tailed recognition benchmarks validate the effectiveness of our method and show our method achieves state-of-the-art performance.

\ifCLASSOPTIONcaptionsoff
  \newpage
\fi

\bibliographystyle{IEEEtran}
\bibliography{egbib}

\end{document}